\documentclass[pdflatex,sn-mathphys]{sn-jnl}

\jyear{2021}%

\theoremstyle{thmstyleone}%
\theoremstyle{thmstyletwo}%

\theoremstyle{thmstylethree}%

\begin{document}

\title[Wavelet-Attention CNN for Image Classification]{Wavelet-Attention CNN for Image Classification}


\author{\fnm{Xiangyu} \sur{Zhao}}\email{imagzhaoxy@njust.edu.cn}

\affil{\orgdiv{School of Computer Science and Engineering}, \orgname{Nanjing University of Science and Technology}, \orgaddress{\street{Xiaolingwei Street}, \city{Nanjing}, \postcode{210094}, \state{Jiangsu}, \country{China}}}

\abstract{The feature learning methods based on convolutional neural network (CNN) have successfully produced tremendous achievements in image classification tasks. However, the inherent noise and some other factors may weaken the effectiveness of the convolutional feature statistics. In this paper, we investigate Discrete Wavelet Transform (DWT) in the frequency domain and design a new Wavelet-Attention (WA) block to only implement attention in the high-frequency domain. Based on this, we propose a Wavelet-Attention convolutional neural network (WA-CNN) for image classification. Specifically, WA-CNN decomposes the feature maps into low-frequency and high-frequency components for storing the structures of the basic objects, as well as the detailed information and noise, respectively. Then, the WA block is leveraged to capture the detailed information in the high-frequency domain with different attention factors but reserves the basic object structures in the low-frequency domain. Experimental results on CIFAR-10 and CIFAR-100 datasets show that our proposed WA-CNN achieves significant improvements in classification accuracy compared to other related networks. Specifically, based on MobileNetV2 backbones, WA-CNN achieves 1.26\% Top-1 accuracy improvement on the CIFAR-10 benchmark and 1.54\% Top-1 accuracy improvement on the CIFAR-100 benchmark.}



\maketitle
\section{Introduction}\label{sec1}
How to effectively extract feature information from images is a challenge in computer vision. Due to the excellent feature extraction ability, Convolutional Neural Network (CNN) has been widely used in computer vision, making a substantial breakthrough in the performance of visual tasks such as image classification, object detection, semantic segmentation and so on~\cite{deng2009imagenet,krizhevsky2009learning,shu2016computational,kumar2017f,hu2018relation,zhang2018context}. Early studies mainly investigate three important aspects of deep networks, namely \textit{depth}, \textit{width}, and \textit{cardinality}. By increasing and improving the network structure, the feature extraction ability of CNN is significantly enhanced~\cite{simonyan2014very,szegedy2015going,he2016deep}.

However, some noise information is residual and some key information is lost in the propagation process. The attention mechanism is a feature weight redistribution strategy, which has achieved excellent results in natural language processing and computer vision~\cite{ba2014multiple,jaderberg2015spatial,vaswani2017attention}. Some researchers add the attention mechanism to CNN in order to improve the ability, which is capturing key feature information. At present, according to different types of features, attention mechanism methods are mainly divided into two categories, i.e., spatial attention and channel attention~\cite{wang2018non,hu_squeeze-and-excitation_2018}. However, using attention may affect the transmission of main feature information to a certain extent.

From the perspective of frequency domain analysis, image information mainly includes low-frequency information and high-frequency information. Low-frequency information mainly consists of image texture and semantic information, and high-frequency information mainly consists of edge and pixel sharp change. Previous work~\cite{wang2020high} validates that humans tend to rely on low-frequent signals in recognizing objects, but CNN mainly perceives the mixed information of low frequency and high frequency. In the training process, high-frequency information has a greater impact on data fitting after many epochs of training, which may lead to overfitting. The high-frequency information in images is mainly divided into two parts: the detailed information related to the data distribution and the noise information. Current methods can not filter out the noise in the high-frequency information completely, only retain the detail of high-frequency information.

Discrete Wavelet Transform (DWT) can obtain high-quality down-sampling information in the field of image processing, which can greatly reduce the loss of down-sampling information in CNN. For example, Li \textit{et al}.~\cite{li2020wavelet} attempted to explore the combination of DWT and CNN to improve the representation ability. Inspired by this, we design a new Wavelet-Attention (WA) block based on DWT and embed WA into CNN, called Wavelet-Attention CNN (WA-CNN). Specifically, WA-CNN decomposes the feature maps into low-frequency components and high-frequency components via the down-sampling operation of DWT. The low-frequency component stores the main information structure of feature maps. These high-frequency components store the detailed information of feature maps while retaining a large amount of noise information. Subsequently, the WA block only captures the detailed information of feature maps in the high-frequency component, and the main information of feature maps in the low-frequency component is not affected. Finally, we conduct experiments on CIFAR-10 and CIFAR-100~\cite{krizhevsky2009learning} datasets to evaluate the effectiveness of the proposed WA-CNN.

Overall, the contributions of this works can be summarized in the following:
\begin{itemize}
  \item We design a new Wavelet-Attention (WA) block that only captures the key information in the high-frequency domain, where the main information in the low-frequency domain is not affected.
  \item We propose a novel Wavelet-Attention CNN (WA-CNN) based on WA block, which accurately captures the detailed information of image data through the WA block, enhancing the feature learning ability of CNN for image classification.
  \item By conducting experiments on two popular benchmarks, i.e., CIFAR-10 and CIFAR-100, the proposed method effectively improves the image classification accuracy compared with the other related methods.
\end{itemize}

The rest of this paper is organized as follows. In Section \ref{sec2}, we summarize the development of CNN, Attention Mechanism, and DWT respectively. In Section \ref{sec3}, we introduce Non-local block and DWT in detail. The new approach is clarified in Section \ref{sec4} and validated experiments are recorded in Section \ref{sec5}. Section \ref{sec6} is the conclusion.

\section{Related Work}\label{sec2}
\textbf{Convolutional Neural Network (CNN).} Based on the powerful representation learning capability, CNN has been widely used in many tasks~\cite{tang2019coherence,8840975,shu2020host,yan2020higcin,yan2021position,kumar2018esumm,kumar2021text} in computer vision. As one of the first convolutional network models, LeNet~\cite{lecun1998gradient} has made a breakthrough in image recognition with its pioneering model design. In 2012, the excellent performance of AlexNet~\cite{krizhevsky2012imagenet} on large-scale image data demonstrated the strong learning capability of deep CNN. After that, VGG~\cite{simonyan2014very} increases network depth by stacking convolution layers, and Inception Net~\cite{szegedy2015going} not only increases the number of convolutional layers but also introduce special multi-branches. However, considering that a naive increase in depth may result in gradient propagation difficulty, ResNet~\cite{he2016deep} solves the optimization problem of deep networks through identity skip-connection. Compared to bloated network structure, the lightweight networks represented by MobileNetV2~\cite{sandler2018mobilenetv2} have also achieved good performance. To date, most general backbone networks mainly target on three basic factors \textit{depth}~\cite{krizhevsky2012imagenet,simonyan2014very,szegedy2015going,he2016deep}, \textit{width}~\cite{szegedy2015going,zagoruyko2016wide,sandler2018mobilenetv2}, and \textit{cardinality}~\cite{xie2017aggregated,chollet2017xception,sandler2018mobilenetv2}. Except for these three basic factors, we focus on the popular attention mechanism.
\vspace{0.2cm}

\noindent\textbf{Attention Mechanism.} When we are observing a scene or object, the human visual system can quickly scan global information and effectively capture and focus on key areas through a series of glimpses~\cite{larochelle2010learning}. Inspired by this, many works~\cite{wang2018non,hu_squeeze-and-excitation_2018,woo_cbam_2018,cao2019gcnet,wang2020eca,shu2021spatiotemporal} introduce the attention mechanism to guide models to learn more critical information and reduce redundancy. NLNet~\cite{wang2018non} uses self-attention to model the pixel-level pairwise relations which construct dense spatial feature maps. Based on the NLNet, various models such as Efficient Attention~\cite{shen2018efficient}, $A^2$-Net~\cite{chen20182} and CCNet~\cite{huang2019ccnet} improve NLNet through different ways. Unlike spatial attention, channel attention models the global context features. SENet~\cite{hu_squeeze-and-excitation_2018} extracts channel features via Global Average Pooling (GAP) and recalibrates the channel features according to importance. ECANet~\cite{wang2020eca} proposes a one-dimensional convolutional layer to replace the full connect layer that uses channel reduction in SENet. Up to now, many works~\cite{park2018bam,woo_cbam_2018,fu2019dual,cao2019gcnet} have created a combination of spatial attention and channel attention. For instance, a lightweight and general module named CBAM is proposed in~\cite{woo_cbam_2018}, which sequentially infers attention maps along two separate dimensions, channel and spatial. GCNet~\cite{cao2019gcnet} combines the advantages of NL and SE modules to design a more effective global context module, obtaining compelling results on assorted tasks. FcaNet\cite{qin2020fcanet} points out that GAP is a special case of Discrete Cosine Transform (DCT) and proposes a multi-spectral channel attention framework, which is an attempt to integrate frequency domain analysis and attention mechanism. By contrast, we explore the application of Discrete Wavelet Transform (DWT) on attention mechanisms.
\vspace{0.2cm}

\noindent\textbf{Discrete Wavelet Transform.} As a common image processing tool of frequency domain analysis, Discrete Wavelet Transform (DWT) is popularly applied in image compression, image denoising, and anti-aliasing~\cite{2003Region,2005The}. Conveniently, DWT can be easily converted to convolution form to be compatible with deep CNN~\cite{huang2017wavelet,savareh2019wavelet}. Up to now, many works~\cite{duan2017sar,liu2018multi,williams2018wavelet,yoo2019photorealistic,li2020wavelet} have researched the fusion of DWT and CNN. Duan \textit{et al}.~\cite{duan2017sar} apply dual-tree complex wavelet transform (DT-CWT) to design a Convolutional-Wavelet Neural Network (CWNN), which can suppress the noise of features and keep their structures. Liu \textit{et al}.~\cite{liu2018multi} propose a Multi-level Wavelet CNN (MW-CNN), which integrates DWT into CNN for image restoration. For decreasing the loss of features, Li \textit{et al}.~\cite{li2020wavelet} replace down-sampling operations with the low-frequency components of DWT and invent a combination of low-frequency components and normal convolution to replace stride convolution. Different from these applications of DWT, our approach aims to enhance the ability to capture detailed information with the attention mechanism.

\section{Analysis on Method}\label{sec3}
In Section \ref{subsec301}, we begin with a brief review of NLNet~\cite{wang2018non}, and concisely analyze the mechanism of spatial attention. Then, the principle of Discrete Wavelet Transform (DWT) and its matrix implementation are introduced in detail in Section \ref{subsec302}.

\subsection{Revisiting the Non-local Block}\label{subsec301}
\begin{figure*}[t]
\centering
\includegraphics[width=\textwidth]{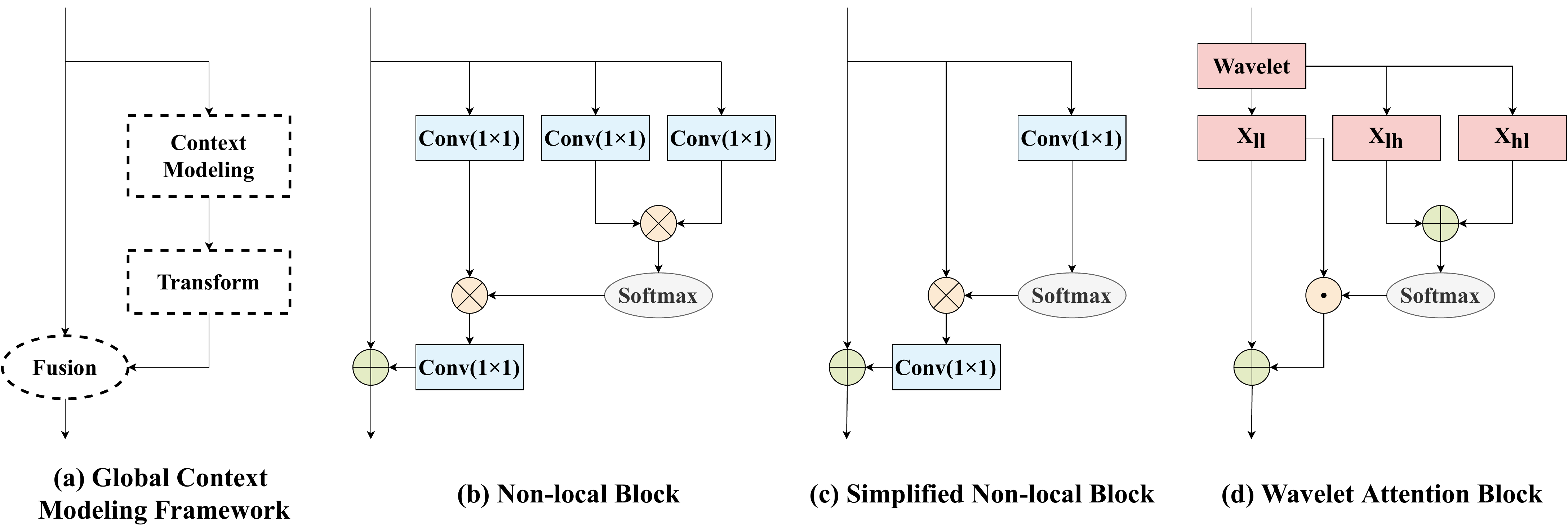}
\caption{Several different attention architectures and our block. $\otimes$ denotes matrix multiplication, $\oplus$ denotes broadcast element-wise addition, and $\odot$ denotes broadcast element-wise multiplication.}\label{fig:2}
\end{figure*}

Inspired by the non-local mean algorithm~\cite{buades2005non}, the non-local block~\cite{wang2018non} can enhance the information response when calculating a certain location feature, by aggregating information between locations. As shown in Figure \ref{fig:2}(b), we denote the feature map input $\mathbf{x}=\{x_{i}\}^{N_p}_{i=1}$, the output of non-local block $\mathbf{z}$ can be expressed as

{
\setlength{\abovedisplayskip}{3pt}
\setlength{\belowdisplayskip}{3pt}
\begin{equation}\label{equ:nonlocal}
\mathbf{z}_{i} = \mathbf{x}_{i} + W_{z} \sum_{j=1}^{N_{p}}\frac{f(\mathbf{x}_{i},\mathbf{x}_{j})}{\mathcal{C}(\mathbf{x})}(W_{v}\cdot\mathbf{x}_{j})
\end{equation}}

\noindent where $i$ represents the index of query positions, and $j\in N_{p}$ indicates all positions in the feature map, $N_{p}=H\times W$ is the number of positions in the feature map, $W_{z}$ and $W_{v}$ represent linear transform matrices, which are implemented as $1\times 1$ convolution. The function $f(\mathbf{x}_{i},\mathbf{x}_{j})$ is used to calculate the relationship between $x_{i}$ and $x_{j}$, and the factor $\mathcal{C}(\mathbf{x})$ is introduced for normalization. In order to meet the requirements of practical application, several forms of $f(\mathbf{x}_{i},\mathbf{x}_{j})$ are designed, as shown in Table \ref{tab1}.

\begin{table}[ht]
\begin{center}
\begin{minipage}{240pt}
\caption{Four different forms of $f(\mathbf{x}_{i},\mathbf{x}_{j})$}\label{tab1}
\centering
\begin{tabular}{ll}
\toprule
Nomenclature & Equation \\
\midrule
Gaussian &  $f(\mathbf{x}_{i}, \mathbf{x}_{j})=e^{\mathbf{x}_{i}^{T} \mathbf{x}_{j}}$ \\
Embedded Gaussian &  $f(\mathbf{x}_{i}, \mathbf{x}_{j})=e^{\theta(\mathbf{x}_{i})^{T} \phi(\mathbf{x}_{j})}$ \\
Dot product &  $f(\mathbf{x}_{i}, \mathbf{x}_{j})=\theta(\mathbf{x}_{i})^{T} \phi(\mathbf{x}_{j})$ \\
Concat &  $f(\mathbf{x}_{i}, \mathbf{x}_{j})=\operatorname{ReLU}(\mathbf{w}_{f}^{T}[\theta(\mathbf{x}_{i}), \phi(\mathbf{x}_{j})])$ \\
\bottomrule
\end{tabular}
\end{minipage}
\end{center}
\end{table}

For example, based on the Embedded Gaussian version of $f(\mathbf{x}_{i},\mathbf{x}_{j})$, the normalized pairwise relationship between position i and j can be expressed as

{
\setlength{\abovedisplayskip}{3pt}
\setlength{\belowdisplayskip}{3pt}
\begin{equation}\label{equ:ij}
\frac{f(\mathbf{x}_{i},\mathbf{x}_{j})}{\mathcal{C}(\mathbf{x})} = \frac{\exp(\left\langle W_{q}\mathbf{x}_{i}, W_{k}\mathbf{x}_{j}\right\rangle)}{\sum_{m=1}^{N_{p}}\exp(\left\langle W_{q}\mathbf{x}_{i}, W_{k}\mathbf{x}_{m}\right\rangle)}
\end{equation}}

According to Equation \ref{equ:nonlocal}, the time complexity of computing attention maps for each query position in the non-local block is $O(N_{P}^{2})$, that resulting in considerable calculation costs. For clarity, GCNet~\cite{cao2019gcnet} visualizes the relationship between different query positions and their generated attention maps, and as a result, their attention maps are almost the same. Theoretically, a non-local block obtains the global context specific to each query position, but the global context after training is not affected by the query position. As shown in the structure in Figure \ref{fig:2}(c), GCNet simplifies the non-local block by sharing a query-independent (global) attention map for all query positions, based on Equation \ref{equ:ij}. Thus, this simplified module can be defined as

{
\setlength{\abovedisplayskip}{3pt}
\setlength{\belowdisplayskip}{3pt}
\begin{equation}\label{equ:gc}
\mathbf{z}_{i} = \mathbf{x}_{i} + W_{v}\sum_{j=1}^{N_{p}}
\frac{\exp(W_{k}\mathbf{x}_{j})}{\sum_{m=1}^{N_{p}}\exp(W_{k}\mathbf{x}_{m})}\mathbf{x}_{j}
\end{equation}}

\noindent where $W_{k}$ and $W_{v}$ denote linear transform matrices. Compare Equation \ref{equ:gc} and Equation \ref{equ:nonlocal}, it can be found that the second term in Equation \ref{equ:gc} is decoupled from the query position $i$, and $W_{v}$ is moved outside to further reduce the computational cost. The global attention map is the weighted average of all location features, which is aggregated to the features of each query location via addition. And some experiments in \cite{cao2019gcnet} validate that the simplified non-local block has comparable performance as the non-local block with significantly lower FLOPs.

\subsection{Discrete Wavelet Transform}\label{subsec302}

\begin{figure}[t]
\begin{center}
\begin{minipage}{0.9\textwidth}
\includegraphics[width=\textwidth]{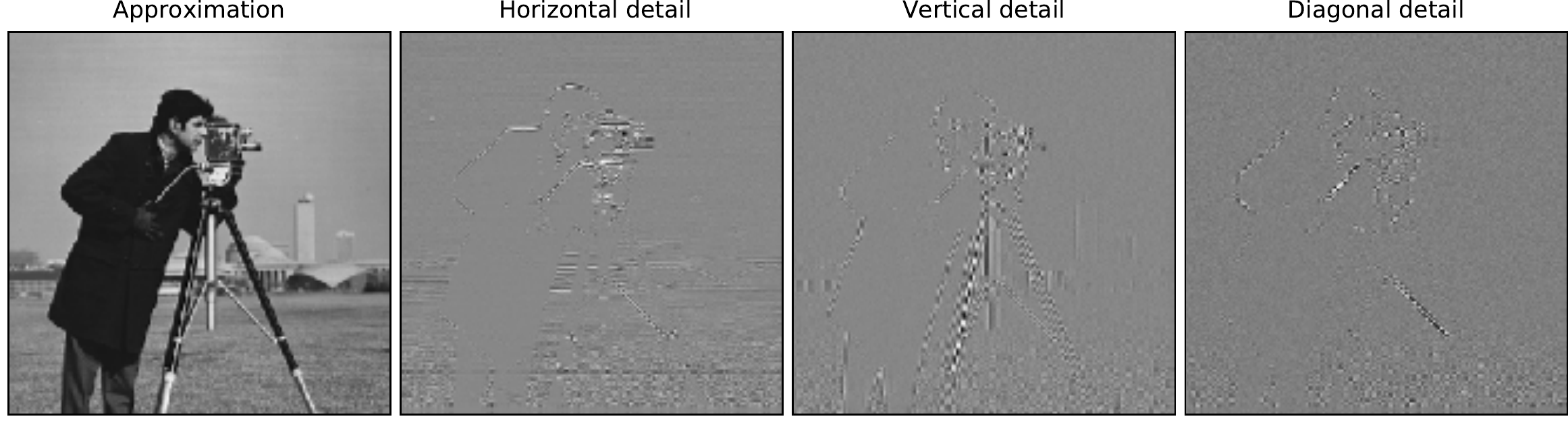}
\caption{An example of image transformation via wavelet family db5}\label{fig:1}
\end{minipage}    
\end{center}
\end{figure}

Wavelet Transform (WT)~\cite{santoso1996power} is a mathematical method to solve the problem of non-stationary signal decomposition in signal processing. By using a set of orthogonal and rapidly decaying wavelet functions to fit signals. WT can obtain different frequency and time positions according to the adjustment of the scale parameter and the translation parameter. To adapt to the discrete data in image processing, DWT is introduced to decompose data into various components in different frequency intervals. Specifically, the mathematical expression of DWT is shown below

{
\setlength{\abovedisplayskip}{3pt}
\setlength{\belowdisplayskip}{3pt}
\begin{equation}\label{equ:wavemath}
W_{\psi}(a, b)=\frac{1}{\sqrt{a}} \sum_{n} \psi^{*}(\frac{n-b}{a})
\end{equation}}

\noindent where $a$ and $b$ represent scale parameter and translation parameter respectively. $\psi(\cdot)$ is the basic wavelet function and $\psi^{*}(\cdot)$ is the conjugate function of $\psi(\cdot)$. Theoretically, the basic wavelet function in the Hilbert space can be decomposed into two parts, the scale function $\phi(\cdot)$ corresponding to the low-frequency part of the original function, and the wavelet function $\psi(\cdot)$ corresponding to the high-frequency part of the original function. In this way, we get an orthogonal wavelet family.

In image processing, two-dimensional DWT (2D-DWT) can transform the original image into a combination of low-frequency information, horizontal high-frequency information, vertical high-frequency information, and diagonal high-frequency information, as shown in Figure \ref{fig:1}. In this work, we use a standard decomposition of 2D-DWT to process image data, which first uses one-dimensional DWT (1D-DWT) to convert row data of the original image and then transform column data of the transformed image. This method can be easily combined with the convolution layer because it can be expressed in matrix form through mathematical transformation. From \cite{li2020wavelet}, we can obtain that the 2D-DWT of input images $X\in \mathbb{R}^{n\times n}$ as

{
\setlength{\abovedisplayskip}{3pt}
\setlength{\belowdisplayskip}{3pt}
\begin{equation}\label{equ:2ddwt}
\begin{aligned}
\mathbf{X}_{ll} &=\mathbf{LXL}^{\mathrm{T}}, & \mathbf{X}_{lh} &=\mathbf{HXL}^{\mathrm{T}} \\
\mathbf{X}_{hl} &=\mathbf{LXH}^{\mathrm{T}}, & \mathbf{X}_{hh} &=\mathbf{HXH}^{\mathrm{T}}
\end{aligned}
\end{equation}}

\noindent where matrix $\mathbf{L}$ and $\mathbf{H}$ are the cyclic matrix composed of wavelet low-pass filter $\{l_{k}\}_{k\in\mathbb{Z}}$ and high-pass filter $\{h_{k}\}_{k\in\mathbb{Z}}$ respectively, and the size are both $\left\lfloor N/2 \right\rfloor \times N$. Here $\mathbf{L}$ and $\mathbf{H}$ can be expanded into the following form

{
\setlength{\abovedisplayskip}{-1pt}
\setlength{\belowdisplayskip}{3pt}
\begin{equation}
\begin{aligned}
\mathbf{L} &= \left(\begin{array}{cccccc}
\cdots & \cdots & & & & \\
\cdots & l_{0} & l_{1} & \cdots & & \\
& & \cdots & l_{0} & l_{1} & \cdots \\
& & & & \cdots & \cdots
\end{array}\right), &
\mathbf{H} &= \left(\begin{array}{cccccc}
\cdots & \cdots & & & & \\
\cdots & h_{0} & h_{1} & \cdots & & \\
& & \cdots & h_{0} & h_{1} & \cdots \\
& & & & \cdots & \cdots
\end{array}\right)
\end{aligned}
\end{equation}}

Based on this, the DWT is implemented as a DWT layer in PyTorch~\cite{li2020wavelet}, which performs DWT on the multichannel data channel by channel. It should be noted that the selected wavelets must have finite filters to ensure the size of the generated matrices are both $\left\lfloor N/2 \right\rfloor \times N$. For example, the simplest wavelet family is Haar, its low-pass filter is $\{l_{k}\}_{k\in\mathbb{Z}}=\{1/\sqrt{2},1/\sqrt{2}\}$, and high-pass filter is $\{h_{k}\}_{k\in\mathbb{Z}}=\{1/\sqrt{2},-1/\sqrt{2}\}$.

\section{Method}\label{sec4}
In this section, we first introduce the global context modeling framework and then discuss in detail the design of the Wavelet-Attention (WA) block and its implementation in CNN, i.e., the proposed Wavelet-Attention CNN (WA-CNN).

\subsection{Global Context Modeling Framework}\label{subsec401}
From the analysis in Section \ref{subsec301}, it is clear that the attention mechanism can be divided into two parts: (a) global attention map gets the global context features; (b) local features are transformed to be the compatible dimension as a feature map. In research \cite{cao2019gcnet}, the structure of attention block can be abstracted as a global context modeling framework in three steps (shown in Figure \ref{fig:2}(a)), defined as

{
\setlength{\abovedisplayskip}{3pt}
\setlength{\belowdisplayskip}{3pt}
\begin{equation}
\mathbf{z}_{i}=F\Bigg(\mathbf{x}_{i}, \delta\Bigg(\sum_{j=1}^{N_{p}} \alpha_{j} \mathbf{x}_{j}\Bigg)\Bigg)
\end{equation}}

Firstly, in step Context Modeling, the framework groups the features of all positions together to obtain the global context features $\sum_{j=1}^{N_{p}}\alpha_{j}\mathbf{x}_{j}$, where $\alpha_{j}$ is the weight of weighted averaging. Then, the features are transformed to capture dependencies in step Transform. Finally, the context features are aggregated to each location via an aggregation function $F(\cdot,\cdot)$ in step Fusion.

This simplified non-local block conforms to the global context modeling framework. Specifically, it obtains the global context features via $1\times1$ convolution $W_{k}$ and softmax function, and then uses $1\times1$ convolution $W_{v}$ to transform the features. Finally, it applies an additional function to aggregate the global context features to the features at each position.

\subsection{Wavelet-Attention Block}\label{subsec402}
Previous researches~\cite{li2020wavelet,zhang2019making} prove that the down-sampling operation of CNN could seriously damage the features. As a suitable solution, the DWT can not only complete the down-sampling operation with high quality but also extract the detailed information of the feature map. Therefore, we refer to the global context modeling framework and propose the WA block, the structure of which is shown in Figure \ref{fig:2}(d).

The WA block performs DWT on the feature map $\mathbf{X}$ to obtain a low-frequency component $\mathbf{X_{ll}}$ and three high-frequency components $\mathbf{X_{lh}}$, $\mathbf{X_{hl}}$, $\mathbf{X_{hh}}$. The low-frequency component preserves the main information structure of the feature map from being corrupted by down-sampling. These high-frequency components retain a lot of noise while retaining the detailed information of the feature map. To reduce the interference of noise, we choose to discard a high-frequency component $\mathbf{X_{hh}}$. Since the WA block can be defined as

{
\setlength{\abovedisplayskip}{3pt}
\setlength{\belowdisplayskip}{3pt}
\begin{equation}\label{equ:wab}
\mathbf{Z} = F\Bigg(\mathbf{X_{ll}},\delta\Bigg(\mathbf{X_{ll}},S\Bigg(F\Bigg(\mathbf{X_{lh}},\mathbf{X_{hl}}\Bigg)\Bigg)\Bigg)\Bigg)
\end{equation}}

\noindent where $F(\cdot,\cdot)$ represents the feature aggregation function, $S(\cdot)$ denotes the softmax function, and $\delta(\cdot,\cdot)$ is the attention map generation function. The WA block extracts the horizontal feature $\mathbf{X_{lh}}$ and vertical feature $\mathbf{X_{hl}}$ from the feature map and aggregates them by broadcast element-wise addition as a global detail feature. Then the softmax function is used to normalize the global detail feature. The global detail feature and the corresponding elements of the low-frequency component $\mathbf{X_{ll}}$ are used to generate the attention map through broadcast multiplication. Finally, the original low-frequency component $\mathbf{X_{ll}}$ adds the attention map by broadcast element-wise addition as output.

\subsection{Wavelet-Attention CNN}\label{subsec403}

The prominent difference between the WA block and the common attention blocks is the former extra includes the down-sampling operation. Intuitively, assuming that the dimension of the input feature map is $H\times W$, the output of the common attention blocks keeps the same dimension, but the WA block is $H/2\times W/2$. Therefore, we need to pay attention to the insertion position of the WA block when constructing a WA-CNN.

\begin{figure}[t]
\begin{center}
\includegraphics[width=0.7\textwidth]{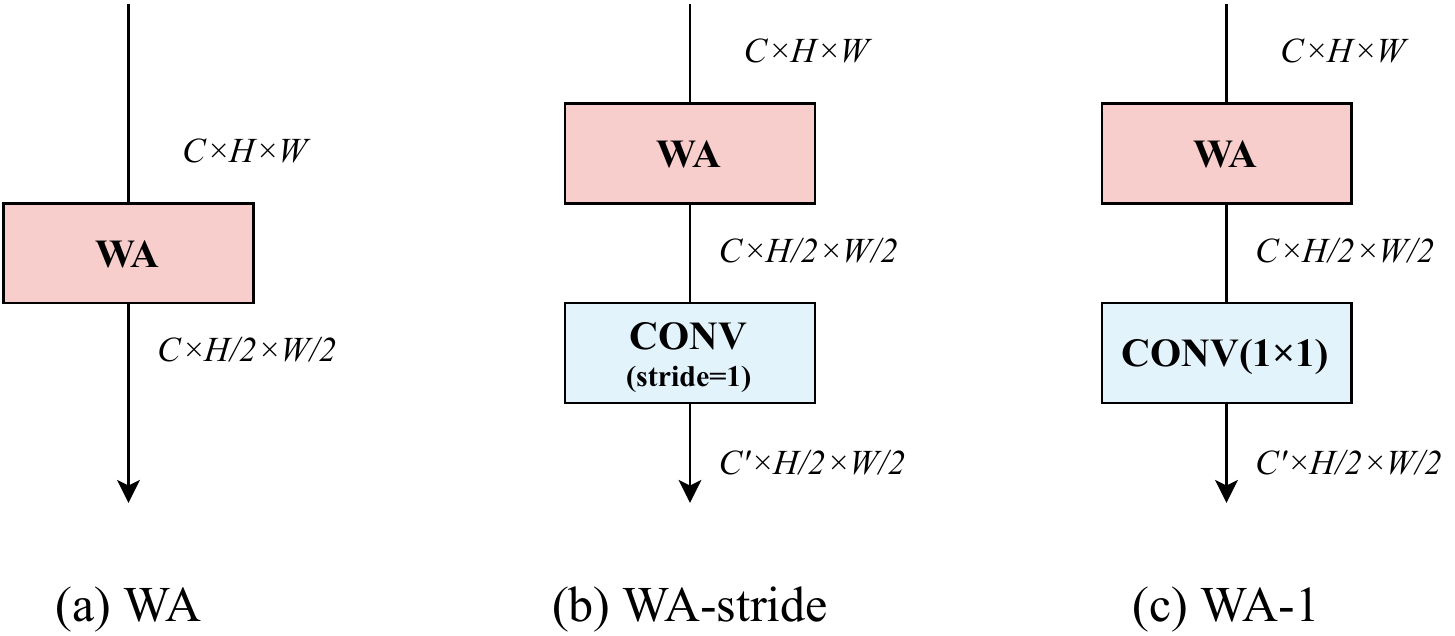}
\caption{Three different variants of WA block. These dimensions of feature maps are $C\times H\times W$.}\label{fig:3}
\end{center}
\end{figure}

Figure \ref{fig:3} shows three different variants of WA block to match the front and back connections at different locations of CNN: (a) WA is an original WA block, which replaces the pooling layer directly. (b) WA-stride is a combination of a WA block and a convolution layer with stride equals one, which replaces the stride convolution with stride equals two. (c) WA-1 is a combination of a WA block and a $1\times 1$ convolution layer, which is used to adjust channel numbers and integrate information across channels.

In practice, the WA block can be easily added to CNN by replacing the max-pooling or average-pooling layer but adding that in other positions requires adjusting network structures. Taking ResNet~\cite{he2016deep} as an example, we elucidate how to embed WA blocks in CNN. Different versions of ResNet are mainly controlled by the number of basic blocks or bottlenecks in sub-layers. Due to the small image size of CIFAR, the structure of ResNet is reduced accordingly. Therefore, the overall structure of ResNet includes three layers. In the second and third layers, ResNet uses $3\times 3$ convolution with a stride of 2 for down-sampling in the first basic block or bottleneck. For the shortcut connection of these subcomponents, ResNet uses the $1\times 1$ convolution to adjust channels and perform down-sampling.

To build a WA-CNN based on ResNet, the WA block needs to be added to basic block and bottleneck. When performing down-sampling, the WA block is activated, and the stride of all convolution layers in this module remains at 1. At the same time, the low-frequency component obtained by DWT is transmitted to the next module by a shortcut connection. This modification applies WA-stride and WA-1 that corresponding to Figure \ref{fig:3}(b) and Figure \ref{fig:3}(c). It is noted that our WA block only acts on the feature extraction stage, and WA-CNN still uses the original classifier in the final stage. Specifically, ResNet uses softmax function to classify the feature maps after average-pooling layer and FC layer, VGG uses softmax function to classify the feature maps after max-pooling layer and FC layers, and MobilenetV2 uses $1\times 1$ convolution to classify the feature maps after average-pooling layer.

\section{Experiments}\label{sec5}

In this section, we conduct a series of experiments on the CIFAR~\cite{krizhevsky2009learning} dataset to evaluate the effectiveness of the WA block. 

\textbf{Overall.} We design three different contrast experiments for the WA block. In the first experiment, the WA block is added in different locations of CNN. And in the second experiment, we test the performance of different wavelet functions based on the same CNN backbone. For the major experiment, we construct the WA-CNN and compare them with several common attention networks to evaluate the performance. All experiments are implemented based on PyTorch and use standard training setting~\cite{yang2021simam}.

\textbf{Dataset.} The CIFAR dataset includes CIFAR-10 and CIFAR-100. CIFAR-10 has 60K $32\times $32 color images in 10 categories, and each category contains 5K train images and 1K test images. CIFAR-100 has 60K $32\times $32 color images in 100 categories, and each category contains 500 train images and 100 test images.

\subsection{Research on the location of adding the WA block}\label{subsec501}

This experiment uses ResNet18 as the benchmark network and the DWT with Haar wavelet. We provide three settings about the position of adding the WA block: layer 2, layer 3, and all layers. The results on CIFAR-10 (C10) and CIFAR-100 (C100) are shown in Table \ref{Ex1}.

The experimental results show that it can improve the classification accuracy of ResNet by adding a WA block to layer 2 and layer 3, respectively. The improvement of a WA block on layer 3 is better than it on layer 2. On the contrary, adding WA blocks to all layers reduce the classification accuracy. One possible explanation is that we transmit the low-frequency component to the next module through the shortcut connection instead of the original transmission, in a basic block where the WA block is activated. For the small spatial size of CIFAR, the impact of one WA block is limited, but the impact of multiple WA blocks is negative. In a word, this experiment verifies the availability of WA-CNN.

\subsection{Research on the different wavelet families}\label{subsec502}

There are many kinds of wavelet families for DWT, and we select three common wavelet families in this experiment. Specifically, "haar" denotes the simplest Haar wavelet, "biorx.y" denotes the Biorthogonal wavelet with orders $(x,y)$, and "dbN" denotes the Daubechies wavelet with approximation order $N$. The length of the wavelet filter increases as the order increases. Haar wavelet and Biorthogonal wavelet have symmetry, while Daubechies wavelet is only approximately symmetrical. We select Daubechies wavelets with shorter filters ("db2" and "db3") because the performance of asymmetric Daubechies wavelet gets worse as the approximation order increases \cite{li2020wavelet}. For this experiment, ResNet34 is used as the backbone to validate five different wavelet functions in three wavelet families. The results of this experiment are recorded in Table \ref{Ex2}.

In Table \ref{Ex2}, the classification accuracy of all wavelet functions exceeds the baseline result. Biorthogonal wavelet (bior2.2) gets the best accuracy on CIFAR-10 and Daubechies wavelet (db3) gets the best accuracy on CIFAR-100. They improve the accuracy of ResNet34 by 0.6\% and 0.67\%, respectively. To sum up, the symmetric wavelets perform better than asymmetric wavelets in image classification.

\begin{table}[t]
\begin{minipage}{\textwidth}
\begin{minipage}{0.45\textwidth}
\caption{The results of WA block be added into different layers. The backbone is ResNet18.}\label{Ex1}
\centering
\begin{tabular}{c|cc}
\hline
\hline
Model & C10 & C100 \\ \hline
Baseline(ResNet18) & 92.44 & 68.64 \\ \hline
layer2 & 92.45 & 68.59 \\
layer3 & \textbf{92.57} & \textbf{69.03} \\
all layer & 92.19 & 68.55 \\
\hline
\hline
\end{tabular}
\end{minipage}
\begin{minipage}{0.55\textwidth}
\caption{Performance comparison of the WA block using different wavelets. The backbone is ResNet34.}\label{Ex2}
\centering
\begin{tabular}{c|c|cc}
\hline
\hline
\multicolumn{2}{c|}{Wavelet} & C10 & C100 \\ \hline
\multicolumn{2}{c|}{Baseline(ResNet34)} & 92.96 & 69.89 \\ \hline
\multicolumn{2}{c|}{Haar} & 93.00 & 70.34 \\ \hline
\multirow{2}{*}{Biorthogonal} & bior2.2 & \textbf{93.56} & 70.33 \\
& bior3.3 & 93.18 & 70.31 \\ \hline
\multirow{2}{*}{Daubechies} & db2 & 93.24 & 70.08 \\
& db3 & 93.19 & \textbf{70.56} \\
\hline
\hline
\end{tabular}
\end{minipage}
\end{minipage}
\end{table}

\subsection{CIFAR Classification}\label{subsec503}
To comprehensively verify the performance of the WA block, we integrate it into several popular CNN for experiments, including VGG~\cite{simonyan2014very}, ResNet~\cite{he2016deep}, and MobileNetV2~\cite{sandler2018mobilenetv2}.

We standardize all experiments with a standard training setting. Regarding data augmentation, we first perform zero-padding with 4 pixels on each side of image data, then randomly cut out $32\times 32$ image from the padded $40\times 40$ image, and finally randomly flip the input image horizontally. For training, we select SGD as the optimizer with the momentum of 0.9, set the batch size to 256, and the weight decay to 0.0005. A total of 160 epochs are trained, and the learning rate is initially set to 0.1, decreased to 0.01 at the 80th epoch and 0.001 at the 120th epoch. All models are trained on a single GPU, and accept the original images for evaluation. To reduce the influence of training fluctuation, we record the average of three training results of each model as the final result.

\begin{figure}[t]
\begin{center}
\includegraphics[width=\textwidth]{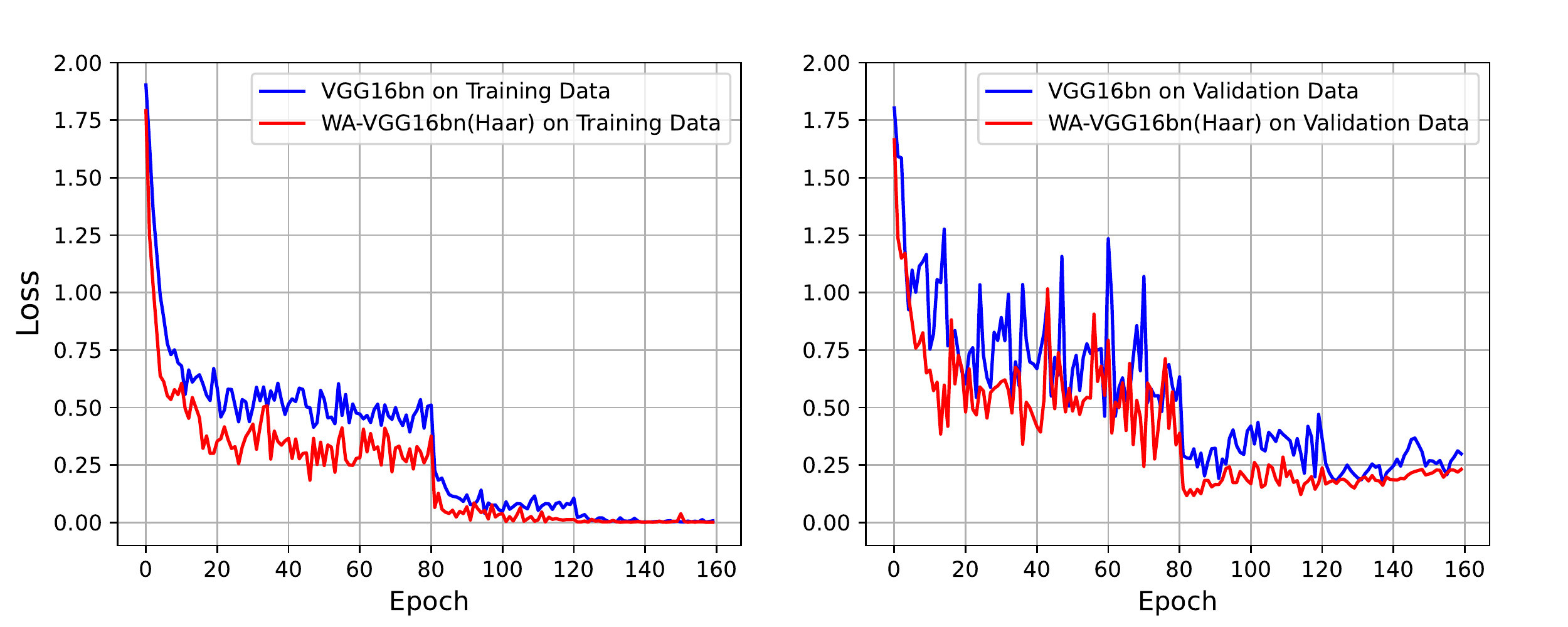}
\caption{The losses of VGG16bn and WA-VGG16bn(Haar) on CIFAR-10.}\label{fig:4}
\end{center}
\vspace{-1em}
\end{figure}

In Figure \ref{fig:4}, we compare the losses of VGG16bn and WA-VGG16bn throughout the training process. Figure \ref{fig:4} adopts blue solid and red solid lines to denote the losses of VGG16bn and WA-VGG16bn, respectively. The training loss of WA-VGG16bn is about 0.2 lower than that of VGG16bn at the first 80 epochs. The right figure shows the losses on validation data. WA-VGG16bn loss is lower than VGG16bn loss in the vast majority of epochs, which leads to the average increase of final classification accuracy by 0.39\%. This expresses that the WA block enhances the training of VGG16bn.

\begin{table}[t]
\centering
\caption{The performance (Top-1 accuracy(\%)) of four common attention methods and our method are tested on CIFAR-10 (C10) and CIFAR-100 (C100) datasets. All results are the average of top-1 accuracy of each model in 3 trials. The best result of each group are displayed in bold black.}\label{Ex3}
\resizebox{\textwidth}{!}{
\begin{tabular}{l|cc|cc|cc|cc|cc}
\hline
\hline
Attention & \multicolumn{2}{c|}{VGG16bn} & \multicolumn{2}{c|}{ResNet18} & \multicolumn{2}{c|}{ResNet34} & \multicolumn{2}{c|}{ResNet50} & \multicolumn{2}{c}{MobileNetV2} \\[3.5pt]
Method & C10 & C100 & C10 & C100 & C10 & C100 & C10 & C100 & C10 & C100 \\[3pt] \hline
Baseline & 93.50 & 73.17 & 92.44 & 68.64 & 92.96 & 69.66 & 93.51 & 71.75 & 91.88 & 71.37 \\[3.5pt] \hline
$+$GC\cite{cao2019gcnet} & 93.77 & 73.49 & 92.50 & 69.24 & 93.27 & 70.14 & 93.58 & 72.57 & 91.75 & 71.86 \\ 
$+$SE\cite{hu_squeeze-and-excitation_2018} & 93.80 & 73.66 & 92.35 & \textbf{69.69} & 92.81 & 70.50 & 93.74 & \textbf{72.89} & 91.82 & 71.54 \\ 
$+$ECA\cite{wang2020eca} & 93.61 & 73.25 & 92.46 & 68.80 & 93.01 & 70.06 & 93.70 & 72.48 & 92.31 & 71.34 \\
$+$CBAM\cite{woo_cbam_2018} & 93.86 & 73.32 & 92.53 & 69.03 & 93.35 & \textbf{71.09} & \textbf{93.80} & 72.51 & 91.98 & 71.89 \\ \hline
$+$WA(Ours) & \textbf{93.89} & \textbf{73.66} & \textbf{92.57} & 69.03 & \textbf{93.56} & 70.56 & 93.63 & 71.90 & \textbf{93.14} & \textbf{72.91} \\
\hline
\hline
\end{tabular}}
\vspace{-2em}
\end{table}

\begin{figure}[t]
\begin{center}
\includegraphics[width=\textwidth]{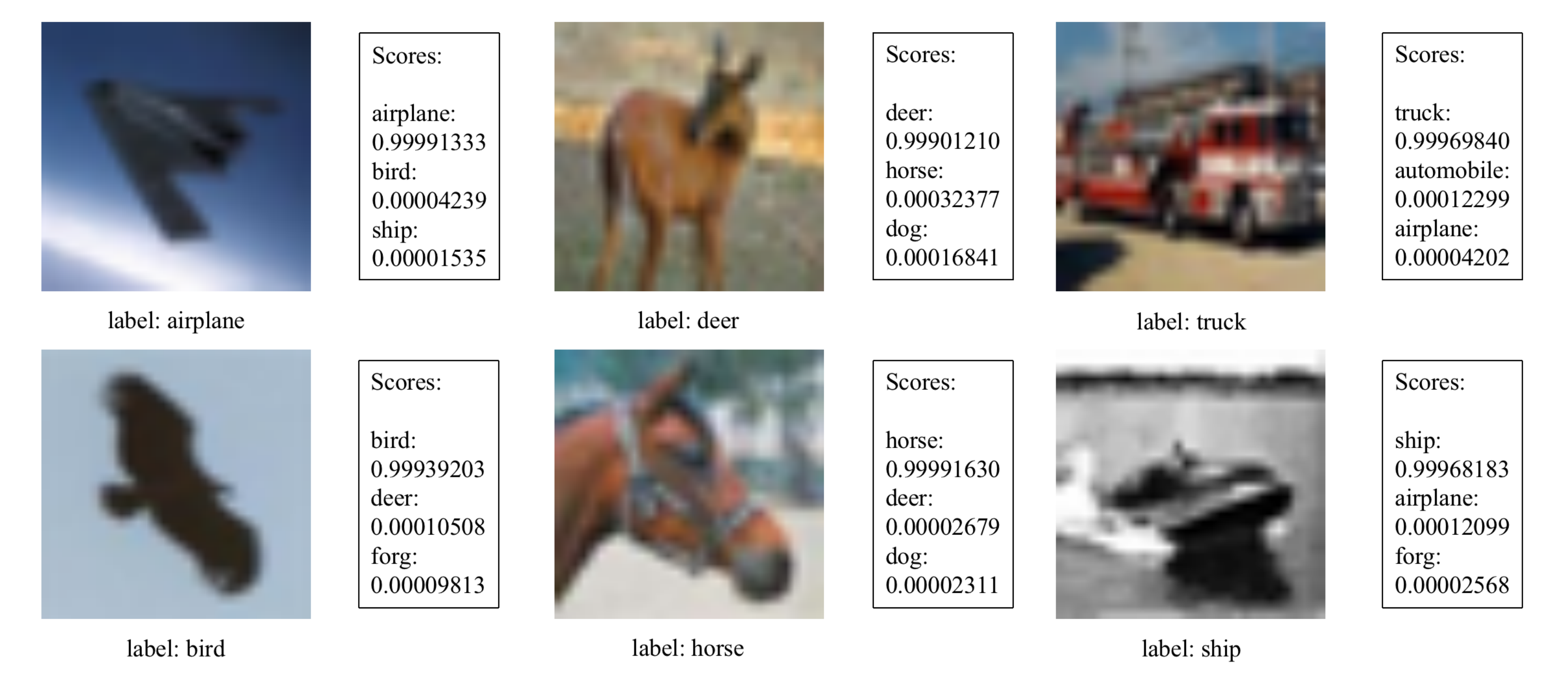}
\caption{The classification results of WA-VGG16bn(bior2.2) on CIFAR-10. The original size of these images is 32 $\times$ 32, so they are blurred.}\label{fig:5}
\end{center}
\end{figure}

From Table \ref{Ex3}, the proposed WA-CNN consistently improves the top-1 accuracy overall baseline networks on both CIFAR-10 and CIFAR-100 datasets. Moreover, the WA block also obtains competitive results against other compared attention methods. In larger networks, our method achieves the best accuracy compared to other modules. Specifically, in VGG16bn, our method achieves the best accuracies on CIFAR-10 (93.89\%) and CIFAR-100 (73.66\%), respectively. Our method improves the accuracy of MobileNetV2 by 1.26\% and 1.54\% on those datasets, respectively. For ResNets, our method also achieves the best accuracy based on ResNet18 and ResNet34 in CIFAR-10. However, our method has no significant improvement compared with other methods in other cases. We have discussed this issue in Section \ref{subsec501}. In Figure \ref{fig:5}, we visualize some classification results from the validation set of CIFAR-10. It can be seen that our method has a strong recognition ability for similar images. On the whole, all these results demonstrate that our WA block is effective for most mainstream networks.

\section{Conclusion}\label{sec6}
In this work, we design a new Wavelet-Attention (WA) block and propose the novel WA-CNN based on it. By using DWT to decompose feature maps, the WA block could guide the model to filter out useless information in the low-frequency domain. Furthermore, the newly designed WA-CNN can integrate the detailed information of features in the high-frequency domain, thus enhancing the feature learning ability of CNN for image classification. The effectiveness of our method is proved by a series of experiments on CIFAR-10 and CIFAR-100. In future work, we will devote more attention to the WA block structure and explore the applications in large-scale datasets.

\backmatter

\bmhead{Acknowledgments}
The work is supported by the National Key R\&D Program of China (No. 2018AAA0102001), the National Natural Science Foundation of China (Grant No. 62072245, and 61932020), the Natural Science Foundation of Jiangsu Province (Grant No. BK20211520).

\bibliography{sn-bibliography}
\end{document}